\documentclass[11pt]{article}
\usepackage{coling2020}
\usepackage{times}
\usepackage{url}
\usepackage{latexsym}

\colingfinalcopy

\let\originalleft\left
\let\originalright\right
\renewcommand{\left}{\mathopen{}\mathclose\bgroup\originalleft}
\renewcommand{\right}{\aftergroup\egroup\originalright}

\usepackage{amsmath}
\usepackage{amssymb}  
\usepackage{bm}  
\usepackage{isomath}

\usepackage{graphicx}
\usepackage{caption}
\usepackage{subcaption}

\title{Contextual BERT: \\Conditioning the Language Model Using a Global State}

\author{Timo I. Denk \\
  Zalando SE \\
  Berlin, Germany \\
  {\tt timo.denk@zalando.de} \\\And
  Ana Peleteiro Ramallo \\
  Zalando SE \\
  Berlin, Germany \\
  {\tt ana.peleteiro.ramallo@zalando.de} \\}

\date{}

\begin{document}

\maketitle

\begin{abstract}
BERT is a popular language model whose main pre-training task is to fill in the blank, i.e., predicting a word that was masked out of a sentence, based on the remaining words.
In some applications, however, having an additional context can help the model make the right prediction, e.g., by taking the domain or the time of writing into account.
This motivates us to advance the BERT architecture by adding a global state for conditioning on a fixed-sized context.
We present our two novel approaches and apply them to an industry use-case, where we complete fashion outfits with missing articles, conditioned on a specific customer.
An experimental comparison to other methods from the literature shows that our methods improve personalization significantly.
\end{abstract}

\section{Introduction}
\label{sec:intro}

Since its publication, the BERT model by \newcite{DBLP:conf/naacl/DevlinCLT19} has enjoyed great popularity in the natural language processing (NLP) community.
To apply the model to a specific problem, it is commonly pre-trained on large amounts of unlabeled data, and subsequently fine-tuned on a target task.
During both stages, the model's only input is a variably-sized sequence of words.

There are use-cases, however, where having an additional context can help the model.
Consider a query intent classifier whose sole input is a user's text query.
Under the assumption that users from different age groups and professions express the same intent in different ways, the classifier would benefit from having access to that user context in addition to the query.
Alternatively, one might consider training multiple models on separate, age group- and profession-specific samples.
However, this approach does not scale well, requires more training data, and does not share knowledge between the models.

To the best of our knowledge, there is a shortcoming in effective methods for conditioning BERT on a fixed-sized context.
Motivated by this, and inspired by the graph-networks perspective on self-attention models \cite{DBLP:journals/corr/abs-1806-01261}, we advance BERT's architecture by adding a global state that enables conditioning.
With our proposed methods [GS] and [GSU], we combine two previously independent streams of work.
The first is centered around the idea of explicitly adding a global state to BERT, albeit without using it for conditioning.
The second is focused on injecting additional knowledge into the BERT model.
By using a global state for conditioning, we enable the application of BERT in a range of use-cases that require the model to make context-based predictions.

We use the outfit completion problem to test the performance of our new methods:
The model predicts fashion items to complete an outfit and has to account for both style coherence and personalization.
For the latter, we condition on a fixed-sized customer representation containing information such as customer age, style preferences, hair color, and body type.
We compare our methods against two others from the literature and observe that ours are able to provide more personalized predictions.

\section{Related Work}
\label{sec:relatedwork}

\paragraph{BERT's Global State}
In the original BERT paper, \newcite{DBLP:conf/naacl/DevlinCLT19} use a \texttt{[CLS]} token which is prepended to the input sequence (e.g., a sentence of natural language).
The assumption is that the model aggregates sentence-wide, global knowledge at the position of the \texttt{[CLS]} token.
This intuition was confirmed through attention score analysis \cite{DBLP:journals/corr/abs-1906-04341}, however, the BERT architecture does not have an inductive bias that aids it.
Recent work therefore treats the \texttt{[CLS]} token differently.
\newcite{DBLP:journals/corr/abs-2007-14062} constrain their BERT variant Big Bird to local attention only, with the exception that every position may always attend to \texttt{[CLS]} regardless of its spatial proximity.

\newcite{Ke2020RethinkingPE} also observe that the \texttt{[CLS]} attention exhibits peculiar patterns.
This motivates them to introduce a separate set of weights for attending to and from \texttt{[CLS]}.
The authors thereby explicitly encode into the architecture that the sequence's first position has a special role and different modality than the other positions.
The result is an increased performance on downstream GLUE tasks.

It is important to note that all related work on BERT's global state does not use the global state for conditioning.
Instead, the architectural changes are solely being introduced to improve the performance on non-contextual NLP benchmarks.

\paragraph{Conditioning on a Context}
To the best of our knowledge, \newcite{DBLP:journals/corr/abs-1812-06705} are the first to provide sentence-wide information to the model to ease the masked language model (MLM) pre-training task.
The authors inject the target label (e.g., positive or negative review) of sentiment data by adding it to the \texttt{[CLS]} token embedding.
In a similar application, \newcite{DBLP:journals/access/LiFXYWJLX20} process the context separately and subsequently combine it with the model output to make a sentiment prediction.

\newcite{DBLP:journals/corr/abs-2004-01881} condition on richer information, namely an \textit{intent}, which can be thought of as a task descriptor given to the model.
The intent is represented in text form, is variably sized, and prepended to the sequence.
This is very similar to a wide range of GPT \cite{radford2019language} applications.

\newcite{DBLP:conf/kdd/ChenHXGGSLPZZ19} condition on a customer's variably-sized click history using a Transformer \cite{DBLP:conf/nips/VaswaniSPUJGKP17}.
The most similar to our work are \newcite{wu2020ssept} who personalize by concatenating every position in the input sequence with a user embedding -- method [C] from Section~\ref{sec:method}.
Their approach, however, lacks an architectural bias that makes the model treat the user embedding as global information.

\paragraph{BERT as a Graph Neural Network (GNN)} 
\newcite{DBLP:journals/corr/abs-1806-01261} introduce a framework that unites several lines of research on GNNs.
In the Appendix, the authors show that -- within their framework -- the Transformer architecture is a type of GNN;
\newcite{joshi2020transformers} supports this finding.
In both cases the observation is that a sentence can be seen as a graph, where words correspond to nodes and the computation of an attention score is the assignment of a weight to an edge between two words.

In the GNN framework, a global state is accessible from every \textit{transfer function} and can be individually updated from layer to layer.\footnote{For details on the GNN definition of a global state we refer the reader to Section~3.2 in \newcite{DBLP:journals/corr/abs-1806-01261}.}
Neither Transformer nor BERT, however, have a global state in that sense.
Inspired by this observation, we introduce a global state and use it for conditioning. 
We explain our two novel methods in the following section, alongside with two that are derived from the literature.

\section{Conditioning BERT With a Global State}
\label{sec:method}

Let $\bm{w}$ denote a sequence of $n$ words $w_i\in\mathbb{V}$ from a fixed-sized vocabulary $\mathbb{V}$.
Further, let $\bm{w}_{-i}$ be the sequence without the $i$th word.
Recall that a vanilla BERT model \cite{DBLP:conf/naacl/DevlinCLT19} can predict the probability
$\Pr(M=w_i\mid \bm{w}_{-i})$
that a word $w_i$ is masked-out in a sequence ($M=w_i$ being the masking event), conditioned on the other words in the sequence.
Next, we introduce four methods to additionally condition BERT on a context vector $\bm{c}\in\mathbb{R}^{d_\text{context}}$, which allows it to predict $\Pr(M=w_i\mid \bm{w}_{-i}, \bm{c})$.

\paragraph{Concat [C]}
Similar to \newcite{wu2020ssept}, we concatenate the context vector with every position in the input sequence.
Let $\bm{x}_i\in\mathbb{R}^{d_\text{model}}$ denote the embedding of word $w_i$ at position $i$.
The resulting input matrix is $\bm{I}_\text{[C]}:=\bm{W}\begin{bmatrix}\bm{x}_1 & \dots & \bm{x}_n\\\bm{c} & \dots & \bm{c}\end{bmatrix}$.
$\bm{W}\in\mathbb{R}^{d_\text{model}\times\left(d_\text{model}+d_\text{context}\right)}$ is a trainable weight matrix that reduces the input dimensionality.

\paragraph{New Position [NP]}
This method adds a new position to the input sequence at which the context is stored.
It is comparable to how \newcite{DBLP:journals/corr/abs-1812-06705} add label information.
Instead of feeding the word sequence $\begin{bmatrix}\bm{x}_1 & \dots & \bm{x}_n\end{bmatrix}$ into the model, we prepend the transformed context $\bm{W}\bm{c}$ to the sequence, where $\bm{W}\in\mathbb{R}^{d_\text{model}\times d_\text{context}}$ is a trainable weight matrix.
The resulting input is $\bm{I}_\text{[NP]}:=\begin{bmatrix}\bm{W}\bm{c} & \bm{x}_1 & \dots & \bm{x}_n\end{bmatrix}$.
The model's attention masks are adjusted, such that every position can attend to the new first position.

\begin{figure}
\centering
\begin{minipage}{.33\textwidth}
  \centering
  \includegraphics[width=.96\linewidth]{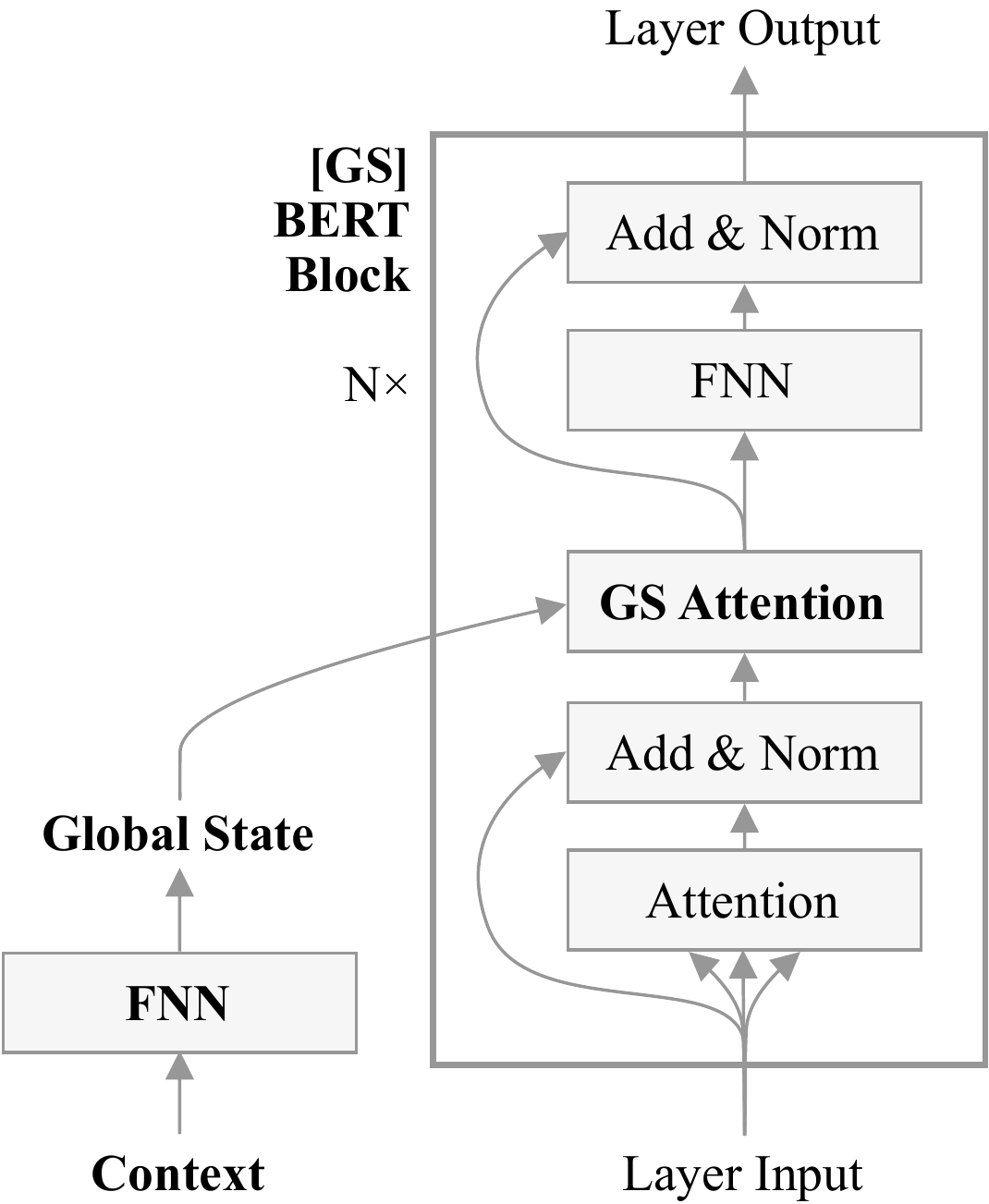}
  \captionof{figure}{A BERT block with global state. Boxes have learned parameters; our additions are bold.}
  \label{fig:tsblock}
\end{minipage}%
\hspace{.5cm}
\begin{minipage}{.63\textwidth}
  \centering
  \includegraphics[width=1\linewidth]{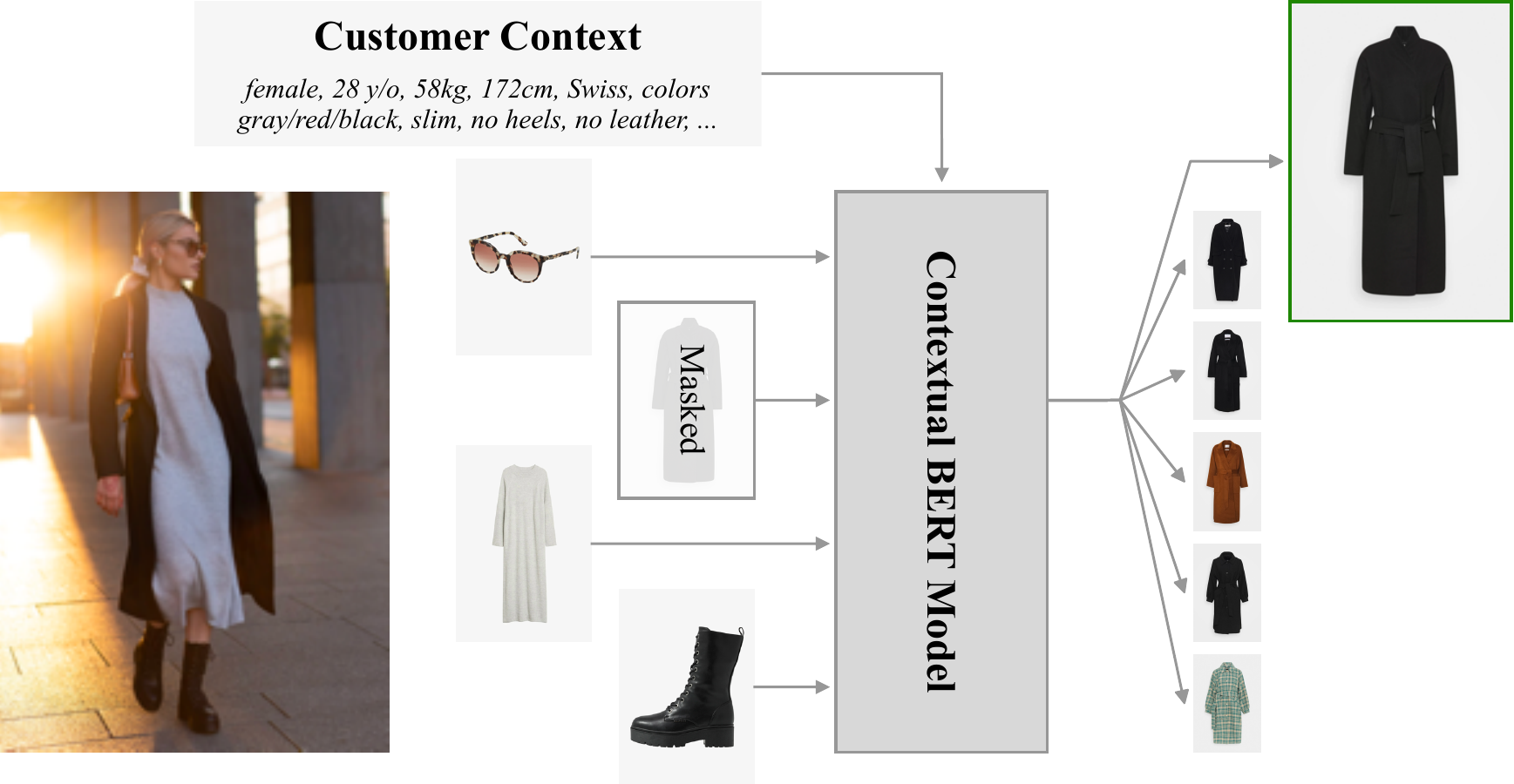}
  \captionof{figure}{The fill-in-the-blank task on fashion outfits. Given a set of articles (left-hand side) and customer context, the model makes several predictions (right-hand side) for a masked out item (here: the coat). The predictions are personalized, because the model is utilizing the customer context.}
  \label{fig:fitbfashion}
\end{minipage}
\end{figure}

\paragraph{Global State [GS]}
Our method is inspired by the GNN perspective on BERT.
Its implementation is similar to the way the Transformer \cite{DBLP:conf/nips/VaswaniSPUJGKP17} decoder attends to the encoder.
[GS] treats the context as a read-only global state from which the internal representations can be updated.
In order to adjust the architecture of BERT accordingly, we insert a \textit{global state attention layer} between the intra-sequence attention and the (originally) subsequent feed-forward neural network (FNN).
Figure \ref{fig:tsblock} shows how the inserted elements fit into the vanilla BERT block.

More formally, let $\bm{X}^{(l)}\in\mathbb{R}^{n\times d_\text{model}}$ be the output of the $l$th BERT block (of which there are $N$); let $\bm{X}^{(0)}:=\begin{bmatrix}\bm{x}_1 & \dots & \bm{x}_n\end{bmatrix}$ be the model input; and let $\bm{\tilde c}:=\operatorname{FNN}(\bm{c})$ be the global state derived from the context vector using a non-linear transformation $\operatorname{FNN}(\bm{x}):=\bm{W}_2\operatorname{max}(0, \bm{W}_1\bm{x}+\bm{b}_1)+\bm{b}_2$.
With our modification, $\bm{X}^{(l)}$ is defined by first performing the normal intra-sequence attention as in BERT
$\bm{A}:=
    \operatorname{Attention}\left(
        \bm{W}^\text{Q}\bm{X}^{(l-1)},
        \bm{W}^\text{K}\bm{X}^{(l-1)},
        \bm{W}^\text{V}\bm{X}^{(l-1)}
    \right)\,;$
multi-head attention can be used here as well. Then, also unchanged,
$\bm{\hat{A}}:=
    \operatorname{LayerNorm}\left(
        \operatorname{Dropout}\left(
            \bm{A}
        \right) + \bm{X}^{(l-1)}
    \right)\,.$
The internal representation is then updated once more by reading\footnote{Note that the global state \textit{sequence} $\begin{bmatrix}\bm{\tilde c}\end{bmatrix}$ does not admit for attention in the sense of selecting a weighted average of multiple vectors, because it consists only of a single vector.
Instead, Equation~\ref{eq:seq2ctxattention} is reduced to the possibility for the $l$th layer to transform the context and update its internal state based on it.} from the global state $\bm{\tilde c}$ with 
\begin{equation}
    \bm{B}:=
        \operatorname{Attention}\left(
            \bm{V}^\text{Q}\bm{\hat{A}},
            \bm{V}^\text{K}\begin{bmatrix}\bm{\tilde c}\end{bmatrix},
            \bm{V}^\text{V}\begin{bmatrix}\bm{\tilde c}\end{bmatrix}
        \right)
        =\bm{V}^\text{V}\begin{bmatrix}\bm{\tilde c}\end{bmatrix}\,,
        \label{eq:seq2ctxattention}\\
\end{equation}
and computing
$\bm{\hat{B}}:=
    \operatorname{LayerNorm}\left(
        \operatorname{Dropout}\left(
            \bm{B}
        \right) + \bm{\hat{A}}
    \right)\,.$
Lastly, the BERT layer output is computed as
$
    \bm{X}^{(l)}:=
        \operatorname{LayerNorm}\left(
            \operatorname{FNN}\left(
                \bm{\hat{B}}
            \right) + \bm{\hat{B}}
        \right)\,.
$
The definitions of $\operatorname{Attention}(\cdot)$, $\operatorname{LayerNorm}(\cdot)$, and $\operatorname{Dropout}(\cdot)$ are identical to \newcite{DBLP:conf/nips/VaswaniSPUJGKP17}.
The weight matrices $\bm{W}$ and $\bm{V}$ are not shared between layers.
The layer indices $^{(l)}$ of $\bm{A}$, $\bm{B}$, and the weight matrices are omitted to aid readability.

\paragraph{Global State With Update [GSU]} Note that in the [GS] method the global state $\bm{\tilde c}$, which is being added to the internal representation in Equation~\ref{eq:seq2ctxattention}, is the same for all blocks.
With [GSU], a global state transfer function
$\bm{\tilde c}^{(l+1)}:=\operatorname{FNN}\left(\bm{\tilde c}^{(l)}\right)$
updates the global state, with its initial value being derived from the context: $\bm{\tilde c}^{(1)}:=\operatorname{FNN}\left(\bm{c}\right)$.
The weights of $\operatorname{FNN}$ are not shared between layers.
Equation~\ref{eq:seq2ctxattention} is updated to read from the global state belonging to the $l$th layer: 
\begin{equation}
    \bm{B}^{(l)}_\text{[GSU]}:=
        \bm{V}^\text{V}\begin{bmatrix}\bm{\tilde c}^{(l)}\end{bmatrix}\,.
\end{equation}

\section{Empirical Evaluation and Discussion}

We evaluate the performance of our proposed methods on a real-world industry problem: personalized fashion outfit completion (see Figure~\ref{fig:fitbfashion}) for Europe's largest fashion platform.
Our proprietary dataset consists of 380k outfits, created by professional stylists for individual customers.
When styling a customer, i.e., putting together an outfit, the stylist has access to all customer features that we later use to condition the model.
Therefore, customer data and outfit are statistically dependent.

The customer features are individually embedded using trainable, randomly initialized embedding spaces.
The per-feature embedding vectors are subsequently concatenated, yielding a context vector of $d_\text{context}=736$.
Features include the customer's age, gender, country, preferred brands/colors/styles, no-go types, clothing sizes, price preferences, and the occasion for which the outfit is needed.
The second model input, namely the outfit itself, is constructed from learned embeddings for every individual fashion article, with $d_\text{model}=128$.
We stack $N=4$ BERT blocks with multi-head attention (eight heads).
For a masked-out item, the model predicts a probability distribution over $\lvert\mathbb{V}\rvert=30\,000$ articles.

While not being an NLP dataset, our data resembles many of the important traits of a textual corpus:
the vocabulary size is comparable to the one of word-piece vocabularies commonly used with BERT models.
Fashion outfits are similar to sentences in that some articles appear often together (match style-wise) and others do not.
Different is the typical sequence length which ranges from four to eight fashion articles, with an average length of exactly five.
In contrast to sentences, outfits do not have an inherent order.
To account for that we remove the positional encoding from BERT so it treats its input as a set.

Table~\ref{tab:mainresults} shows the results of evaluating the four different methods.
We compare cross-entropy and recall@rank (r@$r$ for short) on a randomly selected validation dataset consisting of 17k outfits that are held-out during training.
The r@$r$ is defined as the percentage of cases in which the masked-out item is among the top-$r$ most probable items, according to the model.
Model parameters are counted without embedding spaces for customer features and the last dense layer.

\begin{table}
\begin{center}
\begin{tabular}{lrrrrr}
\hline\relax
\bf Method & \bf Cross-entropy loss & \bf Recall@1 & \bf Recall@5 & \bf Recall@250 & \bf Parameters\\ \hline\relax
[None] & $5.2636\pm 0.0038$ & $8.53\%\pm 0.06$ & $21.44\%\pm 0.04$ & $87.14\%\pm 0.09$ & 546\,432 \\\relax
[C] & $4.9428\pm 0.0047$ & $10.26\%\pm 0.16$ & $25.75\%\pm 0.09$ & $90.76\%\pm 0.10$ & 673\,664 \\\relax
[NP] & $4.9260\pm 0.0078$ & $10.53\%\pm 0.06$ & $26.27\%\pm 0.25$ & $90.80\%\pm 0.11$ & 640\,768 \\\relax
[GS] & $4.8542\pm 0.0084$ & $11.19\%\pm 0.06$ & $27.56\%\pm 0.28$ & $91.35\%\pm 0.13$ & 723\,328 \\\relax
[GSU] & $4.7459\pm 0.0043$ & $12.21\%\pm 0.17$ & $29.40\%\pm 0.06$ & $92.19\%\pm 0.08$ & 921\,856 \\
\hline\relax
\end{tabular}
\caption{\label{tab:mainresults}
Comparison of the different conditioning methods on validation data.
[None] provides no customer context to the model.
Values are means across three training runs reported with standard error.
}
\end{center}
\end{table}

The empirical evaluation reveals the effectiveness of using a context for making predictions.
The model's ability to replicate the stylist behavior better, i.e., achieve a higher r@$r$, improves substantially with the addition of a context.
On r@$1$ we see a relative improvement of $+43\%$ by using the [GSU] method for conditioning over using no customer context at all ([None]) and $+16\%$ compared to [NP] (the best method without global state).

A comparison of the four different conditioning methods shows [GSU] to be most effective, followed by [GS], [NP], and [C].
The methods [C] and [NP] do not have any bias towards treating the context vector specially.
They attend to other positions in the sequence the same way they attend to the context.
The superiority of [GS] and [GSU] can presumably be explained by their explicit architectural ability to retrieve information from the global state and therefore effectively utilize the context for their prediction.

We acknowledge the differences between our outfits dataset and typical NLP benchmarks.
Nonetheless we hypothesize that the effectiveness of our method translates to NLP.
In particular when applied to use-cases in which the modality of context and sequence differ, e.g., for contexts comprised of numerical or categorical meta data about the text.
That is because the model's freedom to read from the context separately allows it to process the different modalities of context and input sequence adequately.

\section{Conclusions and Future Work}

With Contextual BERT, we presented novel ways of conditioning the BERT model.
The strong performance on a real-world use-case provides evidence for the superiority of using a global state to inject context into the Transformer-based architecture.
Our proposal enables the effective conditioning of BERT, potentially leading to improvements in a range of applications where contextual information is relevant.

A promising idea for follow-up work is to allow for information to flow from the sequence to the global state.
Further, it would be desirable to establish a contextual NLP benchmark for the research community to compete on.
This benchmark would task competitors with contextualized NLP problems, e.g., social media platform-dependent text generation or named entity recognition for multiple domains.

\bibliographystyle{coling}
\bibliography{bibliography}

\end{document}